\documentclass{article}
\usepackage{spconf,amsmath,graphicx,hyperref}
\usepackage{array}
\usepackage{graphicx}
\usepackage{booktabs}

\title{Advancing Automated Spatio-Semantic Analysis in Picture Description Using Language Models} 
\name{Si-Ioi Ng$^{1}$, Pranav S. Ambadi$^{1}$, Kimberly D. Mueller$^{2}$, Julie Liss$^{1}$, Visar Berisha$^{1}$}
\address{$^{1}$Arizona State University, USA \\ $^{2}$University of Wisconsin-Madison, USA}
\begin{document}
%
\maketitle
\begin{abstract}
Current methods for automated assessment of cognitive-linguistic impairment via picture description often neglect the visual narrative path - the sequence and locations of elements a speaker described in the picture. Analyses of spatio-semantic features capture this path using content information units (CIUs), but manual tagging or dictionary-based mapping is labor-intensive. This study proposes a BERT-based pipeline, fine tuned with binary cross-entropy and pairwise ranking loss, for automated CIU extraction and ordering from the Cookie Theft picture description. Evaluated by 5-fold cross-validation, it achieves 93\% median precision, 96\% median recall in CIU detection, and 24\% sequence error rates. The proposed method extracts features that exhibit strong Pearson correlations with ground truth, surpassing the dictionary-based baseline in external validation. These features also perform comparably to those derived from manual annotations in evaluating group differences via ANCOVA. The pipeline is shown to effectively characterize visual narrative paths for cognitive impairment assessment, with the implementation and models open-sourced to public \footnote{\url{https://shorturl.at/cmQG3}}. 
\end{abstract}



%
\section{Introduction}
\label{sec:intro}
\begin{figure}[t!]
  \setlength\belowcaptionskip{-0.8\baselineskip}
  \centering
  \includegraphics[width=0.80\linewidth]{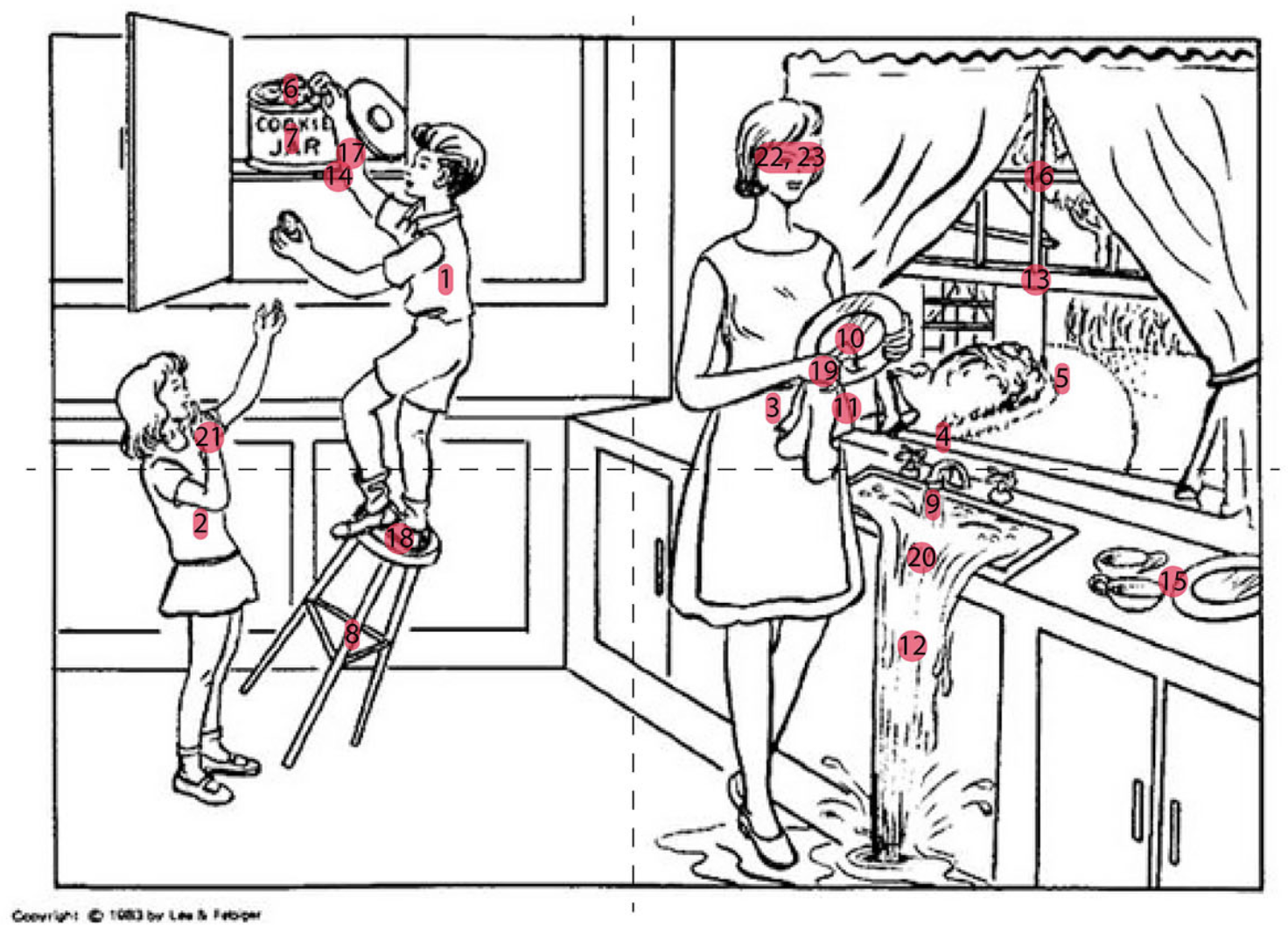}
  \caption{The Cookie Theft picture and CIUs (marked in red).} 
  \label{fig:cookie-theft}
\end{figure}

The picture description task is a widely adopted tool for assessing cognitive and language-specific abilities. It imposes cognitive load on the speakers to amplify the underlying deficits in cognitive functions. Its simplicity in administration and implementation makes it a frequently-employed task in assessing conditions related to cognitive impairments \cite{suh2025picture,  mueller_connected_2018, steel2025visual}. A commonly used stimuli for this task is The Cookie Theft picture \cite{goodglass2001bdae}. It depicts a mother drying dishes, unaware of the overflowing sink. In the background, a boy climbs a stool to reach the jar and steal the cookie, while the a girl stands nearby with an outstretched hand. These objects and actions in the picture can be discretized into content information units (CIUs) to measure the informativeness and relevance of the speaker’s description (see the red marks in Figure \ref{fig:cookie-theft}). 

Recent clinical speech science research has focused on development of models to improve detection of cognitive impairment through picture description \cite{balagopalan20_interspeech, tao2025early, pan2021multi, botelho2024macro}. 
While these existing works focused on leveraging acoustic and linguistic features, Ambadi et. al proposed a graph-theoretic representation to encode CIUs along with their relative spatial position in the picture \cite{ambadi2021spatio}, offering insights into visual processing circuits affected by neurodegenerative changes \cite{jacobs_parietal_2012,salimi_can_2018}. The spatio-semantic features derived from the graph, measuring deficits in visuospatial processing, attentional allocation, and organizational skills, have demonstrated effectiveness in differentiating between healthy controls and cognitively impaired speakers. 

Traditionally, extracting CIUs required labor-intensive manual annotation to build accurate spatio-semantic graphs~\cite{ambadi2021spatio}. To address this, Ng et al. introduced an automated, training-free approach that maps transcripts to CIUs using an expert-curated dictionary~\cite{ng2025automated}, later adopted by Peters et al. for assessing aphasic speech through spatio-semantic features~\cite{peters25_interspeech}. However, the dictionary-based method’s limited vocabulary coverage hinders its ability to handle unseen words, and it fails to account for contextual relationships or interpret sentences holistically during CIU extraction, reducing its effectiveness for diverse or complex datasets.

This study aims to improve the robustness and accuracy of CIU extraction and its spatio-semantic features for cognitive-linguistic analysis, addressing limitations of prior methods \cite{ambadi2021spatio, ng2025automated, peters25_interspeech}. Utilizing a pre-trained BERT language model, our pipeline leverages semantic embeddings to detect diverse CIU expressions and maintain their narrative order in the picture description. We fine tune BERT with a multi-task learning approach, integrating binary cross-entropy for multi-label CIU detection with a pairwise ranking loss to enforce correct sequencing. Model performance is assessed through cross validation, evaluating CIU classification and ordering accuracy. 
External validation and clinical validation are further performed to compare spatio-semantic features derived from the proposed approach against those from the dictionary-based baseline ~\cite{ng2025automated}. 
The trained model and code are openly accessible online, enabling community validation and other applications. 

\begin{figure}[t!]
  \setlength\belowcaptionskip{-0.8\baselineskip}
  \centering
  \includegraphics[width=\linewidth]{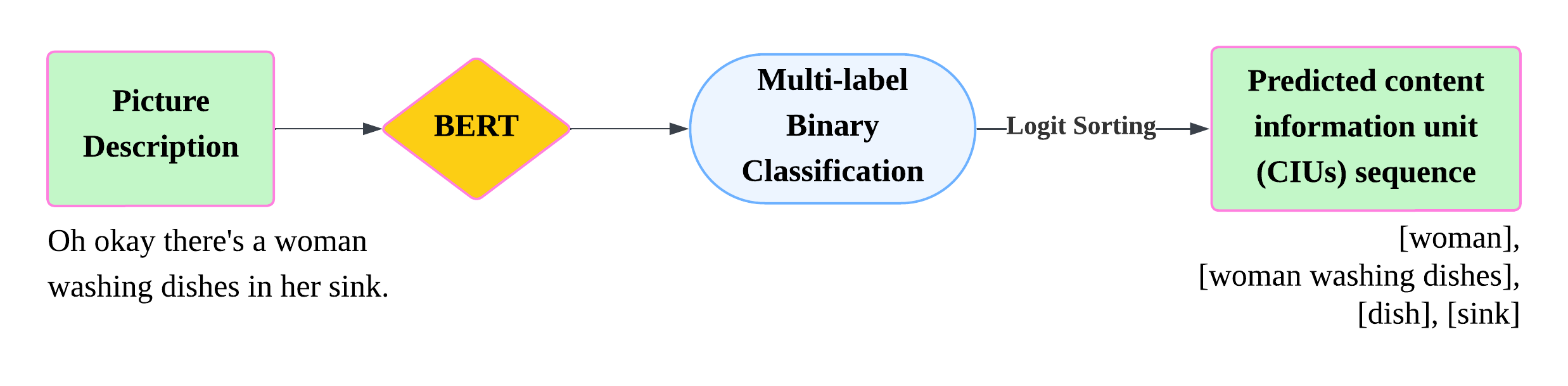}
  \caption{BERT-based CIU extraction workflow from the Cookie Theft picture.} 
  \label{fig:workflow}
\end{figure}

\section{BERT-based CIU Extraction}
The BERT-based pipeline for both CIU \textbf{identification} and \textbf{ordering} is illustrated in Figure \ref{fig:workflow}. 
The input text is first processed through the BERT model to generate contextual embeddings, which are then aggregated via mean pooling. The pooled representation is passed to a linear classification layer that produces logits across the 23 predefined CIU classes. The model identifies predicted CIUs (those with probabilities exceeding 50\%) and orders them based on their logit value to construct the temporal order. 
To fine tune BERT for supporting these dual objectives, we use binary cross-entropy loss as the primary training objective of the CIU extraction model, which performs multi-label classification to detect multiple CIUs within a sentence simultaneously. The loss is given by:  
$$L_{\text{BCE}} = -\frac{1}{K} \sum_{k=1}^{K} \left[ y_k \log(\sigma(s_k)) + (1 - y_k) \log(1 - \sigma(s_k)) \right]$$
where $ K = 23 $ is the number of CIU classes, $ y_k \in \{0, 1\} $ is the ground-truth label for the $ k $-th CIU, $ s_k $ is the logit score for the $ k $-th CIU, and $ \sigma$ is the sigmoid function converting logits to probabilities. 

The secondary objective for the CIU extraction model aims to learn the inherent ordering of CIUs, since the ordering is important for understanding how speakers visually process the Cookie Theft picture. Motivated by similarity learning and margin ranking loss  \cite{chechik2009online, li2019learning, liu2019multi}, we devise an auxiliary pairwise ranking loss alongside the primary binary cross-entropy loss for the multi-label CIU classification. The ranking loss is defined as: 
\
$$L_{\text{rank}} = \frac{1}{N} \sum_{i < j} \max(0, s_j - s_i + m)$$
where $ s_i $ and $ s_j $ are the logit scores for CIUs at positions $ i $ and $ j $ (with $ i < j $). $ m $ is the margin hyperparameter (set to 1 in our experiments), where $ N $ is the number of CIU pairs. 
This loss ensures that earlier CIUs in the ground-truth sequence have higher logit scores than later ones by a set margin, creating a ranking that aligns with the natural narratives. Without it, the model treats the CIUs independently and ignores their sequential dependencies.
During BERT fine tuning, the total loss is a weighted combination: $ L = (1 - \lambda) L_{\text{BCE}} + \lambda L_{\text{rank}} $, with $ \lambda = 0.1 $.

\section{Speech Datasets}
This study utilized speech data from the Wisconsin Registry for Alzheimer’s Prevention (WRAP) dataset \cite{johnson2018wisconsin}, the Wisconsin Alzheimer's Disease Research Center (W-ADRC) dataset \cite{van2021examination}, and the Pitt Corpus from DementiaBank \cite{lanzi2023dementiabank}, which are focused on the Cookie Theft picture description task. 
The WRAP dataset comprises a longitudinal cohorts of  participants, often with familial AD history, who undergo biannual visits for health, lifestyle, and neuropsychological data. 
The Pitt Corpus from DementiaBank includes speech collected from various tasks, including picture description, fluency assessments, story recall, and picture naming. Participants in both datasets are classified as cognitively unimpaired (stable or declining), mild cognitive impairment (MCI), or dementia.
The WRAP dataset and Pitt Corpus were combined for the BERT fine tuning, yielding 2,783 descriptions collected from 1,352 unique speakers. 
The W-ADRC dataset, also compared of a longitudinal mid- to late-life cohorts with similar assessments, provides an additional 256 transcripts from 235 unique speakers for external validation.

All datasets were transcribed in CHAT format \cite{macwhinney2014childes}. 
In each CHAT transcript, CIUs were extracted sentence-wise by trained listeners, with 23 total CIUs in the Cookie Theft image (see Table 1 for the complete list of CIUs).



\section{Experimental Setup}
The BERT-based CIU classifier was fine tuned on the WRAP and Pitt Corpus using the \texttt{bert-base-uncased} pre-trained model on HuggingFace \cite{wolf2019huggingface}, with a hidden size of 768 and 12 transformer layers. 
A dropout rate of 0.2 was applied to mitigate overfitting.  
The fine tuning used 50 epochs with the AdamW optimizer (learning rate 2e-5 for BERT parameters, 1e-3 for the classifier) and combined binary cross-entropy for CIU detection with an auxiliary pairwise ranking loss (margin=1, $\lambda = 0.1$) to enforce CIU ordering. 

We applied 5-fold cross validation, splitting based on speaker groups (avoiding data leakage), to evaluate the accuracy of CIU detection and quality of CIU ordering. CIU detection performance was measured using precision and recall across the 23 CIU categories. The CIU ordering quality was assessed by sequence error rate, which decomposed CIU mismatches into insertions, deletions, and substitutions through computing the Levenshtein distance. The sequence error rate was determined by dividing the total of these discrepancies by the number of actual CIUs. 


The generalization of the BERT model was tested on the W-ADRC dataset, where we compared the Pearson correlations between spatio-semantic features derived from BERT-predicted CIUs and ground-truth CIUs. The clinical effectiveness of features from ground-truth, BERT-predicted, and dictionary-extracted CIUs \cite{ng2025automated} was evaluated using ANCOVA on WRAP and DementiaBank, with spatio-semantic features as dependent variables and age, gender, education level, and unique nodes \cite{ambadi2021spatio} as covariates. The BERT-predicted CIUs were collected from cross-validation evaluation data. The control group comprised 1062 cognitively unimpaired speakers, and the impaired group included 24 speakers with mild cognitive impairment and 189 speakers with dementia. ANCOVA used a significance level of p = 0.05, with F-values indicating differences in feature distributions between groups.

\begin{table}[t]
\centering
\label{precision-recall_table_1}
\footnotesize
\caption{Mean precision and recall per CIU across 5 folds (\%), with standard deviations below 5\% for all CIUs.}
\resizebox{0.85\linewidth}{!}{%
\begin{tabular}{lcc}
\toprule
\textbf{CIU} & \textbf{Prec (\%)} & \textbf{Rec (\%)} \\
\midrule
boy & 95.0 & 98.2 \\
girl & 95.1 & 97.2 \\
woman & 92.9 & 98.3 \\
kitchen & 92.7 & 97.5 \\
outside & 88.3 & 92.2 \\
cookie & 93.7 & 96.0 \\
jar & 96.5 & 96.9 \\
stool & 96.0 & 97.7 \\
sink & 94.8 & 97.3 \\
plate & 90.7 & 95.6 \\
dishcloth & 95.7 & 91.2 \\
water & 96.4 & 98.3 \\
window & 96.4 & 99.2 \\
cupboard & 92.1 & 94.7 \\
dishes & 93.2 & 96.4 \\
curtains & 96.4 & 97.0 \\
boy taking/stealing & 75.7 & 80.1 \\
boy or stool falling & 92.5 & 95.5 \\
woman drying/washing plates & 92.8 & 96.7 \\
water overflowing & 90.6 & 94.1 \\
action performed by girl & 84.7 & 90.4 \\
woman unconcerned by overflowing & 66.4 & 74.6 \\
woman indifferent to the children & 63.8 & 66.1 \\
\bottomrule
\end{tabular}%
}
\end{table}

\section{Experimental Results}
\begin{table*}[th!]
\caption{List of spatio-semantic features and their definition, with Pearson correlation coefficients (r) to ground truth for the proposed method and baseline \cite{ng2025automated}. All correlations are statistically significant (p $<$ 0.05).}
\centering
\resizebox{0.75\linewidth}{!}{
\begin{tabular}{l|l|c|c}
\textbf{Spatio-Semantic Features} & \textbf{Definition} & \textbf{BERT} & \textbf{Dictionary \cite{ng2025automated}} \\ \hline
Avg. X & CIUs' mean X-coordinate & \textbf{0.95} & 0.80 \\ \hline
Std. X & CIUs' standard deviation of X-coordinate & \textbf{0.90} & 0.61 \\ \hline
Avg. Y & CIUs' mean Y-coordinate & \textbf{0.91} & 0.80 \\ \hline
Std. Y & CIUs' standard deviation of Y-coordinate & \textbf{0.93} & 0.79 \\ \hline
Total path distance & Sum of all edge lengths in graph & \textbf{0.97} & 0.85 \\ \hline
Unique nodes & CIUs count without duplicates & \textbf{0.94} & 0.83 \\ \hline
Total path / Unique nodes & \begin{tabular}[c]{@{}l@{}}Total path distance divided by \\ number of unique nodes\end{tabular} & \textbf{0.93} & 0.75 \\ \hline
Nodes & CIUs count With duplicates & \textbf{0.98} & 0.90 \\ \hline
Self cycles & Count of consecutive same CIU & \textbf{0.88} & 0.62 \\ \hline
Cycles & Count of repeated CIUs & \textbf{0.98} & 0.88 \\ \hline
Self cycles (quadrants) & Count of consecutive same quadrant & \textbf{0.92} & 0.81 \\ \hline
Cross ratio (quadrants) & Ratio of inter-quadrant to intra-quadrant edges & \textbf{0.64} & 0.31 \\ \hline
\end{tabular}
}
\label{tab:spatio_semantic_features_correlations}
\end{table*}



\begin{figure}[t!]
  \setlength\belowcaptionskip{-0.8\baselineskip}
  \centering
  \includegraphics[width=0.95\linewidth]{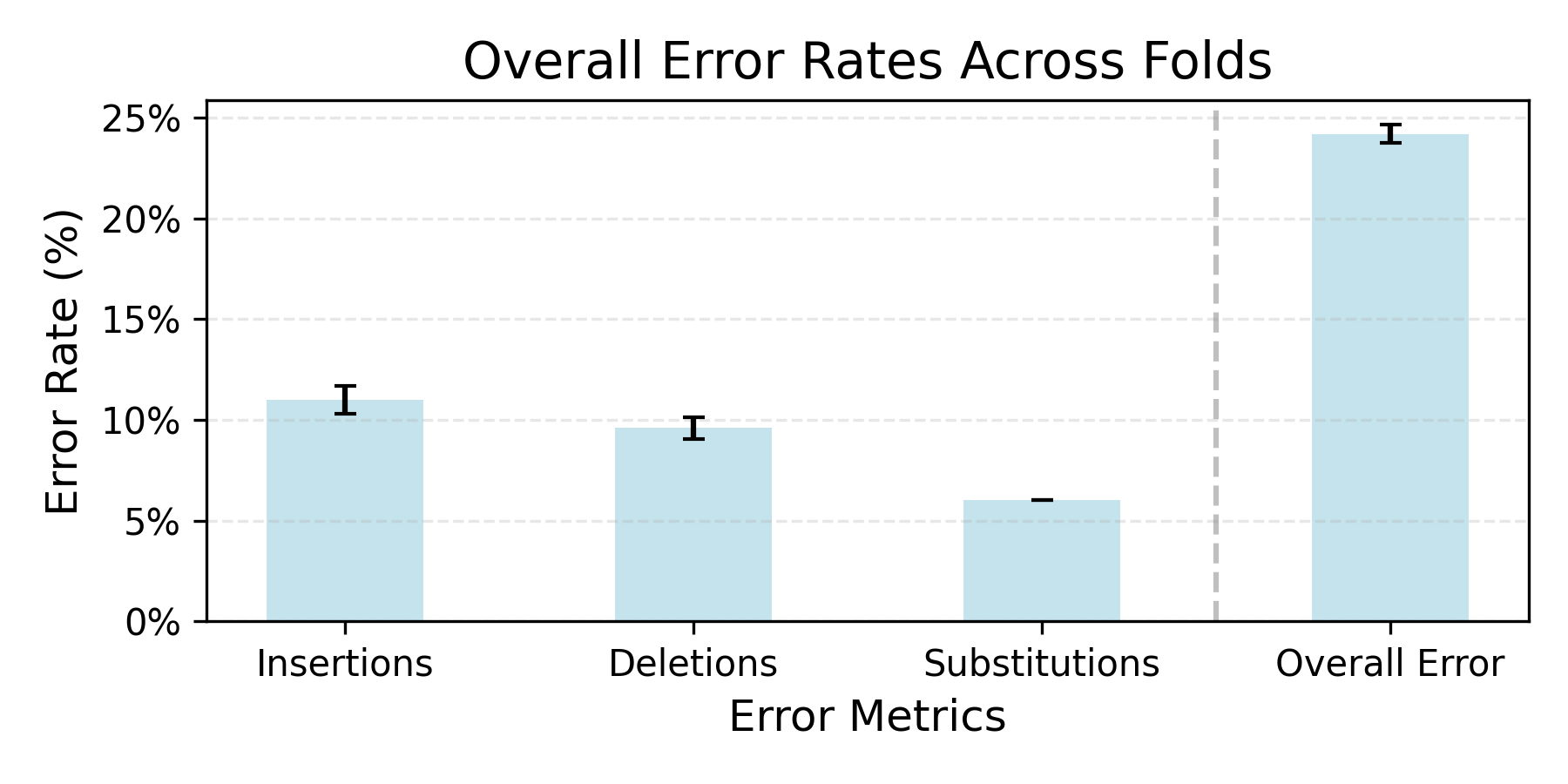}
  \caption{Performance of CIU ordering.} 
  \label{fig:ordering_score}
\end{figure}

Table 1 presents the mean precision and recall for detecting each of the 23 CIUs across five cross-validation folds in the multi-label classification setting, with standard deviations below 5\% for all CIUs. The fine tuned BERT model demonstrates robust detection, with 20 CIUs achieving over 80\% in both precision and recall. 
Recall generally surpasses precision, suggesting the model effectively captures true positives but has a tendency for false positives, such as CIU insertions in predicted sequences. 
However, for \textit{boy taking/stealing}, \textit{woman unconcerned by overflowing}, and \textit{woman indifferent to the children}, precision ranges from 63.8\% to 75.7\% and recall from 66.1\% to 80.1\%, reflecting challenges in detection due to their semantic complexity, dependence on broader context, and lower training data frequency. 

Figure ~\ref{fig:ordering_score} reports speaker-level sequence error rates, with insertion rates of approximately 11\%, deletion rates of 10\%, and substitution rates of 6\%, yielding a consistent overall sequence error rate of 24\% across folds. 
The insertion errors echo with the higher recall rates reported earlier in Table 1, while the substitution errors arise during logit-based sequence sorting. 
These results highlight the BERT model’s ability to accurately detect CIUs and maintain their narrative order, enabling effective derivation of spatio-semantic features for downstream applications. 

Table~\ref{tab:spatio_semantic_features_correlations} reports the Pearson correlation coefficients for spatio-semantic features derived from the BERT-extracted CIUs, evaluated against ground-truth features on the W-ADRC dataset which was excluded from BERT fine tuning. To enhance robustness for the external validation, the BERT model was fine tuned on the full combined dataset from the WRAP and Pitt Corpus. Compared to the dictionary-based baseline \cite{ng2025automated}, the BERT-based approach shows stronger alignment with true spatial (e.g. mean and standard deviation of X/Y coordinates) and sequential patterns (e.g., total path distance, cycle counts), with significantly higher correlations. 
Notable improvement over the dictionary baseline include Std. X (0.90 vs. 0.61), self cycles (0.88 vs. 0.62) and cross-quadrant ratios (0.64 vs. 0.31). We observe that the baseline produces longer sequences with excessive repetitions (e.g. favoring tagging CIUs such as boys and girls and their actions), increasing intra-quadrant transitions relative to ground truth. The sequence differences lead to divergent values in these features since they are particularly sensitive to repetition and transition patterns. Our proposed method mitigates these issues, achieving closer alignment with ground truth by reducing verbosity and balancing quadrant transitions.



\begin{table}[b!]
\caption{ANCOVA test results (* p $<$ 0.05); \textdagger: Unique nodes is used as the dependent variable, not as a covariate.}
\centering
\resizebox{0.95\linewidth}{!}{
\begin{tabular}{l|c|c|c}
\textbf{Spatio-Semantic Features} & \textbf{Ground truth} & \textbf{BERT} & \textbf{Dictionary \cite{ng2025automated}} \\ \hline
Avg. X & 2.82 & 0.50 & 0.33 \\ \hline
Std. X & 2.38 & 1.71 & 6.41* \\ \hline
Avg. Y & 0.25 & 1.22 & 1.61 \\ \hline
Std. Y & 0.03 & 0.11 & 0.14 \\ \hline
Total path distance & 21.78* & 27.16* & 32.99* \\ \hline
\textdagger Unique nodes & 31.67* & 25.80* & 29.08* \\ \hline
Total path / Unique nodes & 25.80* & 30.71* & 32.41* \\ \hline
Nodes & 32.7* & 23.60* & 43.03* \\ \hline
Self cycles & 3.50 & 1.26 & 1.91 \\ \hline
Cycles & 34.75* & 23.60* & 43.74* \\ \hline
Self cycles (quadrants) & 4.98 & 9.21* & 50.66* \\ \hline
Cross ratio (quadrants) & 0.70 & 0.76 & 8.21* \\ \hline\hline
\textbf{Mean F-value} & \textbf{13.45} & \textbf{12.14} & \textbf{20.88} \\ \hline
\textbf{Std. F-value} & \textbf{14.35} & \textbf{12.10} & \textbf{18.99} \\ \hline
\end{tabular}
}
\label{tab:spatio_semantic_features_fvalues}
\end{table}

Table \ref{tab:spatio_semantic_features_fvalues} presents the ANCOVA test results for spatio-semantic features derived from ground-truth, BERT-based, and dictionary-based CIUs \cite{ng2025automated}, using the combined WRAP and Pitt Corpus. Features such as total path distance, unique nodes, total path / unique nodes, nodes, and cycles consistently showed significant F-values across all methods, effectively distinguishing cognitively unimpaired from impaired groups. 
BERT-based spatio-semantic features yield F-values (mean 12.14, s.d. 12.10), closely aligned with ground-truth values (mean 13.45, s.d. 14.35). This indicates strong similarity in distinguishing clinical classes, whereas dictionary-based features (mean 20.88, s.d. 18.99) exhibit greater variability.
Notably, the dictionary approach over-tags repetitive CIUs in impaired speakers, inflating same-quadrant counts that triggers statistical significance and large F-value in self cycles (quadrants). BERT-based approach shows similar but less pronounced inflation, yielding a marginal F-value increase. 
The dictionary's over-tagging reduces cross-quadrant transitions and slightly increases variability in Std. X. This triggers the significance in both. 
Overall, BERT’s alignment with ground truth ensures more reliable and consistent spatio-semantic feature extraction. 
The dictionary approach, while simpler, remains a practical alternative but is less precise due to its variability.

\section{Conclusion}
This study presents BERT-based pipeline for extracting and ordering Content Information Units (CIUs) from picture description. By fine tuning the BERT with a loss function combining binary cross-entropy for CIU detection with an auxiliary pairwise ranking loss, we achieved high accuracy and effective sequence reconstruction. 5-fold cross-validation showed precision and recall scores above 80\% in detecting various CIUs with sequence error rate of 24\%, confirming the model’s consistent performance across varied speaker subsets. Compared to a dictionary-based baseline, our approach better aligns with ground-truth spatio-semantic features, as shown by Pearson correlation coefficients in the external validation. Clinical validation further confirms that spatio-semantic features derived from BERT-extracted CIUs perform comparably to those from manually annotated CIUs in ANCOVA tests, that assess group differences between healthy and cognitively impaired speakers. Future work will explore applying spatio-semantic features to other neurodegenerative disorders for broader clinical generalizability. 

\section{Acknowledgment}
This work was supported in part by the National Institute on Aging under Research Grant R01-AG082052.


\bibliographystyle{IEEEbib}
\bibliography{strings,refs}

\end{document}